\documentclass[final]{cvpr}

\usepackage{times}
\usepackage{epsfig}
\usepackage{graphicx}
\usepackage{animate}
\usepackage{amsmath}
\usepackage{amssymb}
\usepackage{subfigure}
\usepackage{marvosym}

\usepackage[pagebackref=true,breaklinks=true,colorlinks,bookmarks=false]{hyperref}

\def\cvprPaperID{5092}

\def\confYear{2021}
\begin{document}

\title{MotionRNN: A Flexible Model for Video Prediction with\\Spacetime-Varying Motions}

\author{
  Haixu Wu\thanks{Equal contribution}, Zhiyu Yao\footnotemark[1], Jianmin Wang, Mingsheng Long (\Letter) \\
  School of Software, BNRist, Tsinghua University, China \\
  {\tt\small \{whx20,yaozy19\}@mails.tsinghua.edu.cn, \{jimwang,mingsheng\}@tsinghua.edu.cn} 
}

\maketitle

\begin{abstract}
  This paper tackles video prediction from a new dimension of predicting spacetime-varying motions that are incessantly changing across both space and time. Prior methods mainly capture the temporal state transitions but overlook the complex spatiotemporal variations of the motion itself, making them difficult to adapt to ever-changing motions. We observe that physical world motions can be decomposed into \emph{transient variation} and \emph{motion trend}, while the latter can be regarded as the accumulation of previous motions. Thus, simultaneously capturing the transient variation and the motion trend is the key to make spacetime-varying motions more predictable. Based on these observations, we propose the \emph{MotionRNN} framework, which can capture the complex variations within motions and adapt to spacetime-varying scenarios. MotionRNN has two main contributions. The first is that we design the \emph{MotionGRU} unit, which can model the transient variation and motion trend in a unified way. The second is that we apply the MotionGRU to RNN-based predictive models and indicate a new flexible video prediction architecture with a \emph{Motion Highway}, which can significantly improve the ability to predict changeable motions and avoid motion vanishing for stacked multiple-layer predictive models. With high flexibility, this framework can adapt to a series of models for deterministic spatiotemporal prediction. Our MotionRNN can yield significant improvements on three challenging benchmarks for video prediction with spacetime-varying motions.
\end{abstract}

\vspace{-10pt}
\section{Introduction}

Real-world motions are extraordinarily complicated and are always varying in both space and time. It is extremely challenging to accurately predict motions with space-time variations, such as the
deformation, accumulation, or dissipation of radar echoes
in precipitation forecasting. Recent advanced deterministic video prediction models, such as PredRNN \cite{wang2017predrnn}, MIM \cite{WangZZLWY19} and Conv-TT-LSTM \cite{su2020convolutional} mainly focus on capturing the simple state transitions across time. They overlook the complex variations within the motions so that they cannot predict accurately under the highly changing scenario. Besides, optical-flow based methods \cite{shi2017deep, reda2018sdc} use local-invariant state transitions to capture the short-term temporal dependency but lack the characterization of long-term motion trends. These methods may degenerate significantly when modeling ever-changing motions.

\begin{figure}[tb]
  \centering
  \includegraphics[width=\columnwidth]{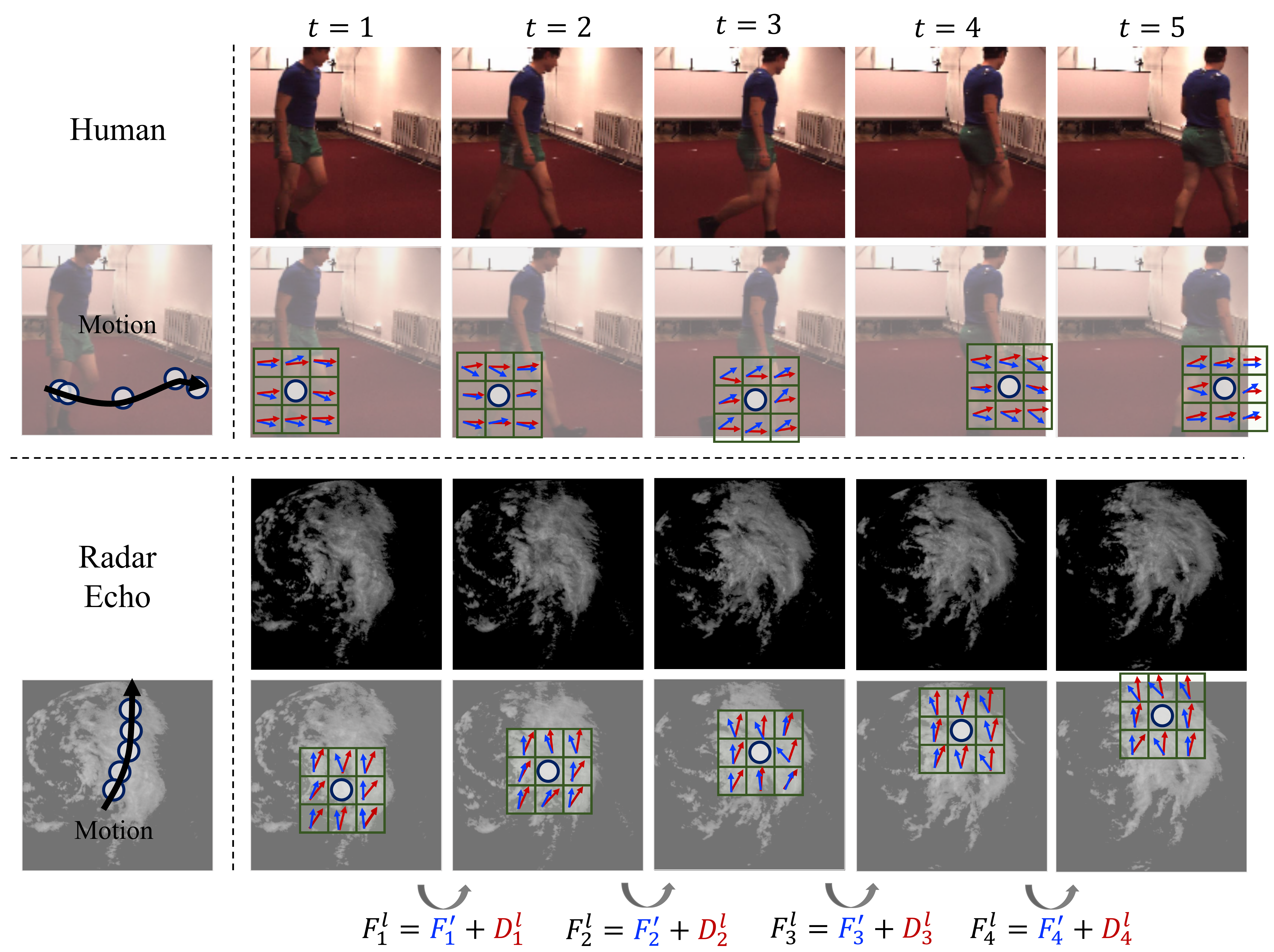}
  \caption{Two cases of real-world spacetime-varying motions. The movements $\mathcal{F}_{t}^l$ (shown in \textbf{black} arrows) of human legs or radar echoes can be decomposed into \emph{transient variation} and \emph{motion trend}.
  Our MotionRNN captures the transient variation $\mathcal{F}_{t}^\prime$ (\textcolor{blue}{blue} arrows) and the motion trend $\mathcal{D}_{t}^l$ (\textcolor{red}{red} arrows) simultaneously.}
  \label{fig:human_show}
  \vspace{-10pt}
\end{figure}

We observe that physical world motions can be naturally decomposed into the \textit{transient variation} and \textit{motion trend}. The transient variation can be seen as the deformation, dissipation, speed change or other variations of each local region instantly. As shown in Figure \ref{fig:human_show}, when a person is running, different parts of the body will have various transient movement changes across time, \textit{e.g.} the left and the right legs are taken forward alternately. Moreover, the natural spatiotemporal processes are following the rule of the trend, especially for physical motions. In the running scenario of Figure \ref{fig:human_show}, the body sways up and down at each time step, but the man keeps moving forward from left to right following the unchanging tendency. The motion follows the characteristics behind the physical world in a video sequence, such as inertia for objects, meteorology for radar echoes, or other physical laws, which can be seen as the \textit{motion trend} of the video. Considering the decomposition of the motion, we should capture the transient variation and the motion trend for better space-time varying motion prediction. 

We go beyond the previous state-of-the-art methods for deterministic spatiotemporal prediction \cite{wang2017predrnn,WangZZLWY19,su2020convolutional} and propose a novel \textbf{MotionRNN} framework. To enable more expressive modeling of the spacetime-varying motions, MotionRNN adapts a \textbf{MotionGRU} unit for high-dimensional hidden-state transitions, which is specifically designed to capture the transient variation and the motion trend respectively. Inspired by the residual shortcuts in the ResNet \cite{He2015Deep}, we improve the \textbf{Motion Highway} across layers within our framework to prevent the captured motions from vanishing and provide useful contextual spatiotemporal information for the MotionRNN. Our MotionRNN is flexible and can be easily adapted to the existing predictive models. Besides, MotionRNN achieves new state-of-the-art performance on three challenging benchmarks: a real-world human motion benchmark, a precipitation nowcasting benchmark, and a synthetic varied flying digits benchmark. The contributions of this paper are summarized as follows:
\begin{itemize}
  \label{question}
    \item {Based on the key observation that the motion can be decomposed to transient variation and the motion trend, we design a new MotionGRU unit, which could capture the transient variation based on the spatiotemporal information and obtain the motion trend from the previous accumulation in a unified way.}
    \item We propose the MotionRNN framework, which unifies the MotionGRU and a new Motion Highway structure to make spacetime-varying motions more predictable and to mitigate the problem of motion vanishing across layers in the existing predictive models.
    \item {Our MotionRNN achieves the new state-of-the-art performance on three challenging benchmarks. And it is flexible to be applied together with a rich family of predictive models to yield consistent improvements. }
\end{itemize}

\section{Related Work}

\subsection{Deterministic Video Prediction}

Recurrent neural networks (RNNs) have been wildly used in the field of video prediction to model the temporal dependencies in the video \cite{Ranzato2014Video, Oh2015Action, srivastava2015unsupervised, denton2018stochastic,  de2016dynamic, liu2017video, shi2017deep, Kalchbrenner2017Video, villegas2017learning, reda2018sdc, guen2020disentangling, yao2020unsupervised,su2020convolutional}.
To learn spatial and temporal content and dynamics in a unified network structure, Shi \textit{et al.} \cite{shi2015convolutional} proposed the convolutional LSTM (ConvLSTM), extending the LSTM with convolutions to maintain spatial information in the sequence model. The fusion of CNNs and LSTMs makes the predictive models capable to capture the spatiotemporal information. Finn \textit{et al.} extended ConvLSTM for robotics to predict the transformation kernel weights between robot states. Wang \textit{et al.} \cite{wang2017predrnn} introduced PredRNN, which makes the memory state update along a zigzag state transition path across stacked recurrent layers using the ST-LSTM cell. For capturing long-term dynamics, E3D-LSTM \cite{wang2019eidetic} incorporated 3D convolution and memory attention into the ST-LSTM, which can capture the long-term video dynamics. Su \textit{et al.} \cite{su2020convolutional} presented a high-order convolution LSTM (Conv-TT-LSTM) to learn the spatiotemporal correlations by combining the history convolutional features. 

Still, previous spatiotemporal predictive models mainly focus on spatiotemporal state transitions but ignore internal motion variations. When it comes to instantly-changing motions, these predictive models may not behave well. To learn the coherence between frames, some video prediction methods are based on the optical flow \cite{sutton1988learning, Simonyan14}. SDC-Net \cite{reda2018sdc} learns the transformation kernel and kernel offsets between frames based on the optical flow. TrajGRU \cite{shi2017deep} also follows the idea of optical flow to learn the receptive area offsets for a special application of precipitation nowcasting. Villegas \textit{et al.} \cite{Villegas2017Decomposing} leveraged the optical flow for short-term dynamic modeling. These optical-flow based methods capture the short-term temporal dynamics effectively. However, they only treat the video as the instantaneous translation of pixels between adjacent frames and may ignore the motion trend of object variations.

Note that these methods are generally based on the RNN, such as LSTMs. In this paper, we propose a flexible external module for RNN-based predictive models without changing their original predictive framework. Unlike previous predictive learning methods, our approach focuses on modeling the within-motion variations, which could learn the explicit transient variation and remember the motion trend in a unified way. Our method naturally complements existing methods for learning spatiotemporal state transitions and can be applied with them for more powerful video prediction. 

\subsection{Stochastic Video Prediction}

In addition to these deterministic video prediction models, some recent literature has explored the spatiotemporal prediction problem by modeling the future uncertainty. These models are based on adversarial training \cite{Mathieu2015Deep,vondrick2016generating,tulyakov2018mocogan} or variational autoencoders (VAEs) \cite{babaeizadeh2017stochastic, tulyakov2018mocogan, denton2018stochastic, lee2018stochastic, villegas2019high, castrejon2019improved, franceschi2020stochastic}.
These stochastic models could partially capture the spatiotemporal uncertainty by estimating the latent distribution for each time step. They did not attempt to explicitly model the motion variation, which is different from our MotionRNN. Again, MotionRNN can be readily applied with these stochastic models by replacing their underlying backbones.

\section{Methods}

Recall our observation as shown in Figure \ref{fig:human_show}: real-world motions can be decomposed into the \emph{transient variation} and \emph{motion trend}. In the spirit of this observation, we propose the flexible MotionRNN framework with a motion highway, which could effectively enhance the ability to adapt to the spacetime-varying motions and avoid the motion vanishing. Further, we propose a specifically designed unit named MotionGRU, which can capture the transient variation and motion trend in a unified recurrent cell. This section will first describe the MotionRNN architecture and illustrate how to adapt MotionRNN to the existing RNN-based predictive models. Next, we will present the unified modeling of transient variation and motion trend in the MotionGRU unit.

\subsection{MotionRNN}

\begin{figure}[htb]
  \centering
  \vspace{-10pt}
  \includegraphics[width=1\columnwidth]{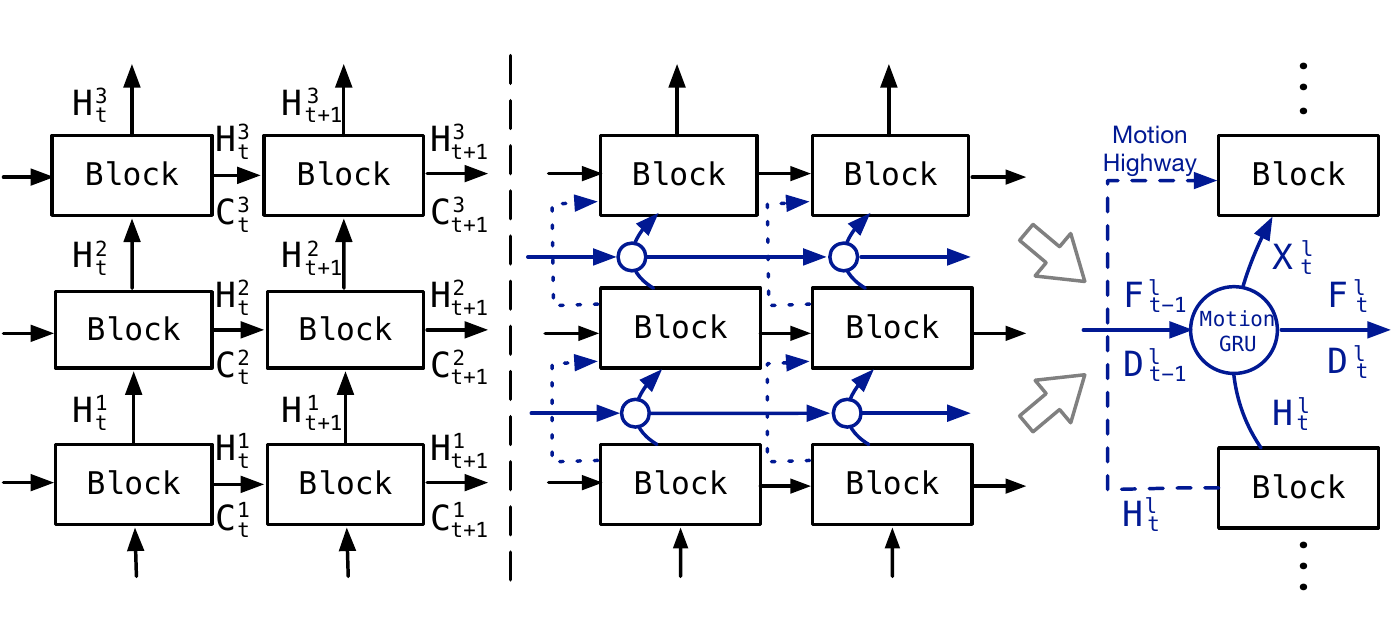}
  \vspace{-20pt}
  \caption{
  An overview of typical architecture of predictive frameworks: RNN-based spatiotemporal predictive networks (\textbf{left}), MotionRNN framework (\textbf{right}) which embeds the MotionGRU (\textcolor{blue}{blue} circles) between layers of the original models. The \textcolor{blue}{blue} dashed lines between stacked layers present the Motion Highway.
  }
  \label{fig:pipeline}
\end{figure}

Typically, RNN-based spatiotemporal predictive models are in the forms of stacked blocks, as shown in Figure \ref{fig:pipeline}. Here we use each block to indicate the predictive RNN unit, such as ConvLSTM \cite{shi2015convolutional} or ST-LSTM \cite{wang2017predrnn}. In this framework, the hidden states transit between predictive blocks and are controlled by the inner recurrent gates. However, when it comes to spacetime-varying motions, the gate-controlled information flow would be overwhelmed by incessantly making quick responses to the transient variations of motions. Besides, it also lacks motion trend modeling.

To tackle the challenge of spacetime-varying motions modeling, the MotionRNN framework incorporates the MotionGRU unit between the stacked layers as an operator without changing the original state transition flow (Figure \ref{fig:pipeline}). MotionGRU can capture the motion and conduct a state transition to the hidden states based on the learned motion. However, we find that motion will blur and even vanish when the transited features pass through multi-layers. Motivated by this observation, MotionRNN introduces the Motion Highway to provide an alternative quick route for the motion context information. We find that Motion Highway could effectively avoid motion blur and constrain the object in the right location from the visualization in Figure \ref{fig:human_mc}.

In detail, the MotionRNN framework inserts the MotionGRU between layers of the original RNN blocks. Take ConvLSTM \cite{shi2015convolutional} as an example. After the first layer, the overall equations for the $l$-th layer at time step $t$ are as follows:
\begin{equation}\label{equ:overall}
  \begin{split}
    \mathcal{X}_{t}^{l}, \mathcal{F}_{t}^{l}, \mathcal{D}_{t}^{l} &= \textrm{MotionGRU}(\mathcal{H}_{t}^{l}, \mathcal{F}_{t-1}^{l}, \mathcal{D}_{t-1}^{l}) \\
    \mathcal{H}_{t}^{l+1}, \mathcal{C}_{t}^{l+1} &= \textrm{Block}(\mathcal{X}_{t}^{l}, \mathcal{H}_{t-1}^{l+1}, \mathcal{C}_{t-1}^{l+1}) \\
    \mathcal{H}_{t}^{l+1} &= \mathcal{H}_{t}^{l+1}+(1-o_{t})\odot\mathcal{H}_{t}^{l}, \\
  \end{split}
\end{equation}
where $l\in \{1, 2,\cdots, L\}$. Tensors $\mathcal{F}_{t}^{l}$ and $\mathcal{D}_{t}^{l}$ denote the transient filter and the trending momentum from MotionGRU respectively. We will give detailed descriptions to MotionGRU in the next section. The input $\mathcal{X}_{t}^{l}$ of the Block has been transited by MotionGRU. $\mathcal{H}_{t-1}^{l+1}$, $\mathcal{C}_{t-1}^{l+1}$ are the hidden state and memory state from the previous time step respectively, which are the same as original predictive blocks. $o_t$ is the output gate of the RNN-based predictive block, which reveals the constantly updated memory in LSTMs.

The last equation presents the motion highway, which compensates the predictive block's output by the previous hidden state $\mathcal{H}_{t}^{l}$. We reuse the output gate to expose the desired unchanging content information. This highway connection provides extra details to the hidden states and balances the invariant part and the changeable motion part.

Note that MotionRNN does not change the state transition flows in the original predictive models. Thus, with this high flexibility, MotionRNN can adapt to a rich family of predictive frameworks, such as ConvLSTM \cite{shi2015convolutional}, PredRNN \cite{wang2017predrnn}, MIM \cite{WangZZLWY19}, E3D-LSTM \cite{wang2019eidetic}, and other RNN-based predictive models. It can significantly enhance spacetime-varying motion modeling of the existing predictive models.

\subsection{MotionGRU}

As mentioned above, towards modeling the spacetime-varying motions, our approach presents the MotionGRU unit to conduct motion-based state transitions by modeling the motion variation. In video prediction, the motion can be presented as pixels displacement corresponding to the hidden states transitions in RNNs. We use the MotionGRU to learn the \emph{pixel offsets} between adjacent states. The learned pixel-wise offsets are denoted by \emph{motion filter} $\mathcal{F}_{t}^l$. Considering that real-world motions are the composition of transient variations and motion trends, we specifically design two modules in the MotionGRU to model these two components respectively (Equation \ref{equ:mornn_unit}). 

\begin{figure}[htb]
\centering
\includegraphics[width=0.8\columnwidth]{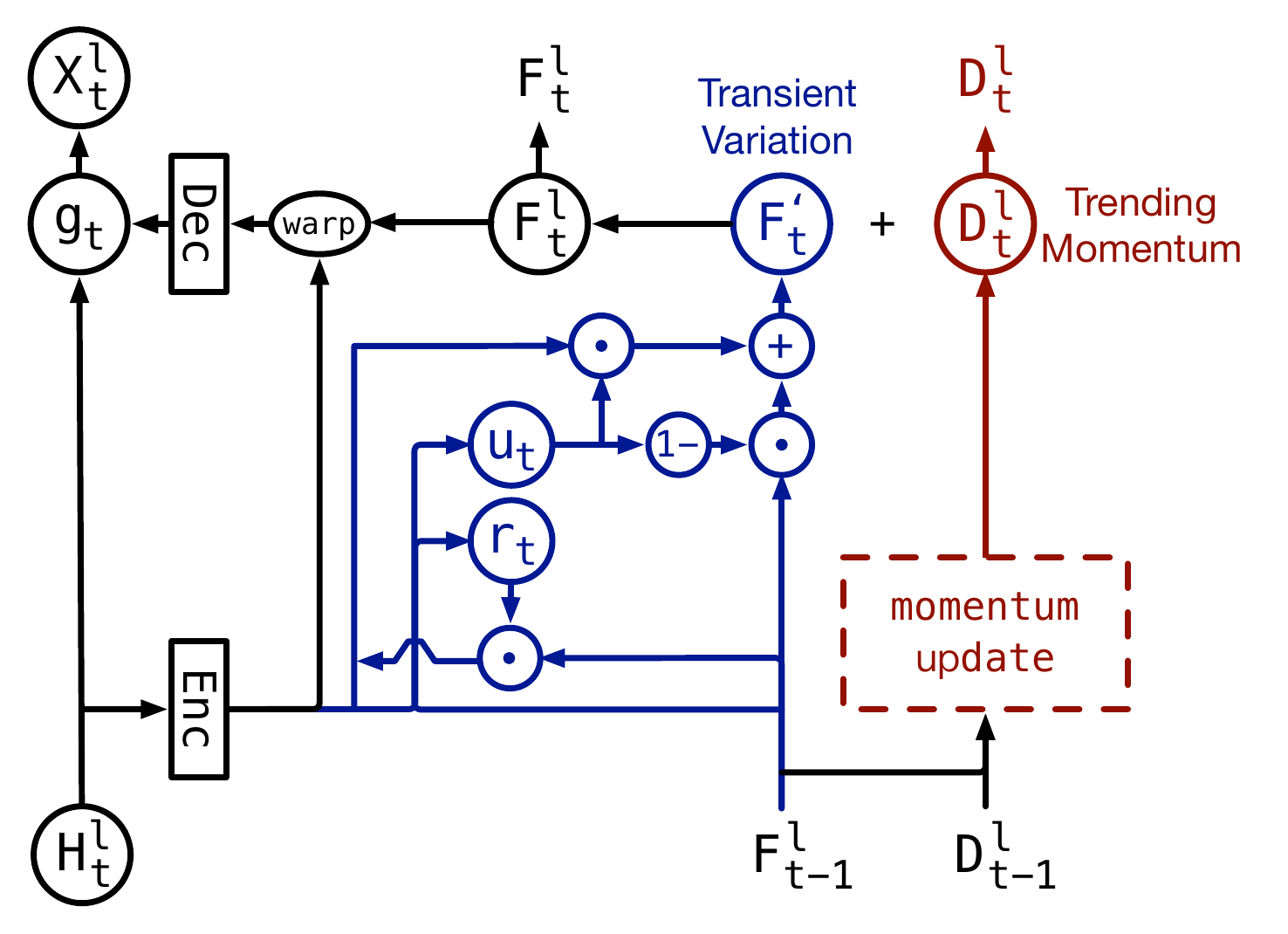}
\caption{MotionGRU unit's architecture. The \textcolor{blue}{blue} part is to capture the transient variation $\mathcal{F}_{t}^\prime$.
The trending momentum $\mathcal{D}_{t}^l$ accumulates the motion tendency in an accumulation way (\textcolor{red}{red} part).}
\vspace{-10pt}
\label{fig:mornn_unit}
\end{figure}

\subsubsection{Transient Variation} 

In a video, the transient variation at each time step is not only based on the spatial context but also presents high temporal coherence. For example, the waving hands of a man follow a nearly continuous arm rotation angle between adjacent frames. Motivated by the spatiotemporal coherence of transient variations, we adapt a ConvGRU \cite{shi2017deep} to learn the transient variation. With this recurrent convolutional network, the learned transient variation could consider the instant states and maintain the spatiotemporal coherence of variations. The equations of the transient-variation learner of the $l$-th MotionGRU at time step $t$ are shown as follows:
\begin{equation}\label{equ:motionGRU_unit}
  \begin{split}
  u_t &= \sigma\left(W_{u}\ast \text{Concat}([\text{Enc}(\mathcal{H}_t^l), \mathcal{F}_{t-1}^l])\right) \\
  r_t &= \sigma\left(W_{r}\ast \text{Concat}([\text{Enc}(\mathcal{H}_t^l), \mathcal{F}_{t-1}^l])\right) \\
  z_t &= \tanh\left(W_{z}\ast \text{Concat}([\text{Enc}(\mathcal{H}_t^l), r_t \odot \mathcal{F}_{t-1}^l])\right) \\
  \mathcal{F}_t^\prime &=  u_t \odot z_t 
   + (1-u_t) \odot \mathcal{F}_{t-1}^l. \\
  \end{split}
\end{equation}
We use $\mathcal{F}_{t}^\prime  = \textrm{Transient}\left(\mathcal{F}_{t-1}^l, \textrm{Enc}(\mathcal{H}_{t}^l)\right)$ to summarize the above equations.
$\sigma$ is the sigmoid function, $W_{u}$, $W_{r}$ and $W_{z}$ denotes the $1\times 1$ convolution kernel, $\ast$ and $\odot$ denote the convolution operator and the Hadamard product respectively. $u_t$ and $r_t$ are the update gate and reset gate in ConvGRU \cite{shi2017deep}, and $z_t$ is the reseted feature of current moment. $\text{Enc}(\mathcal{H}_{t}^l)$ encodes the input from the last predictive block. $\mathcal{F}_{t-1}^l$ presents motion filter from the previous time step for capturing the transient variations. Transient variation $\mathcal{F}_{t}^\prime$ for current frame is calculated with the update gate $u_t$. Note that transient variation $\mathcal{F}_{t}^l$ presents the transition of each pixel's position between adjacent states. Thus, all the gates, $z_{t}$, and $\mathcal{F}_{t}^l$ are in the offset space, which are learned filters and different from spatiotemporal states $\mathcal{H}_{t}^{l},\mathcal{C}_{t}^{l}$.

\subsubsection{Trending Momentum} 

\begin{figure}[tbp]
  \centering
  \includegraphics[width=1\columnwidth]{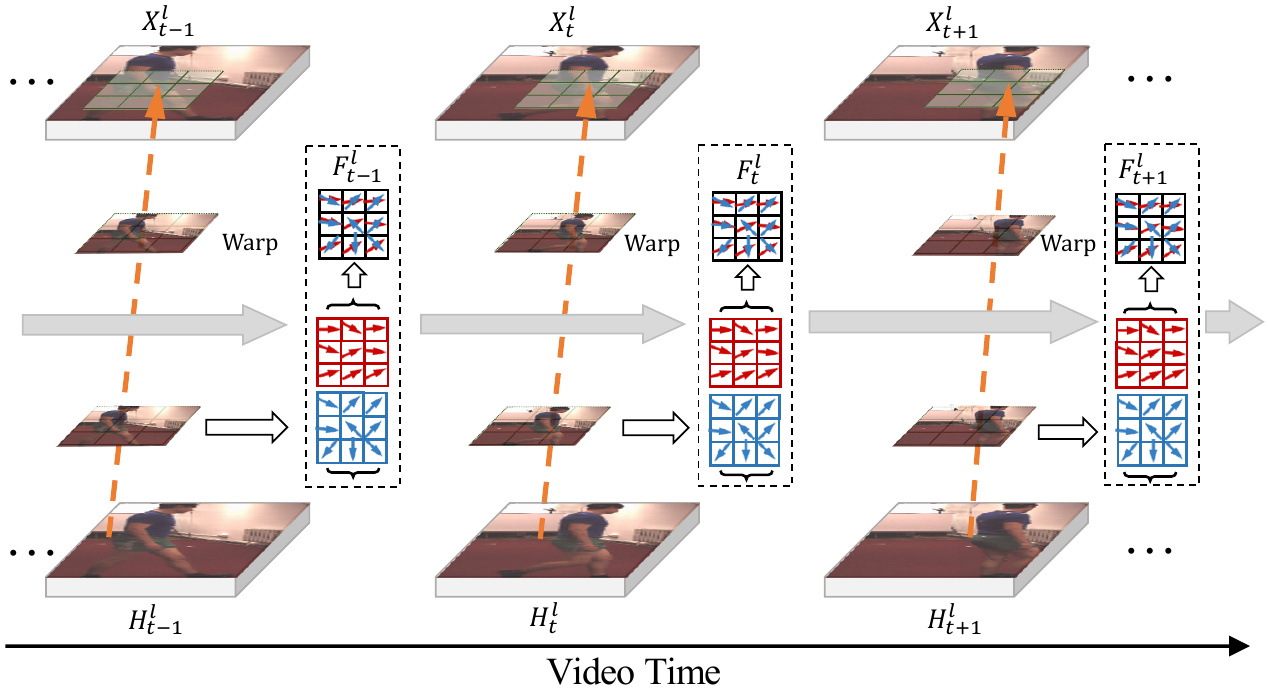}
  \caption{State transitions by MotionGRU. The motion filter $\mathcal{F}_{t}^l$ is combined by the transient variation (\textcolor{blue}{blue} square) and trending momentum (\textcolor{red}{red} square). The new transited state is obtained by the Warp operation based on the learned motion filter.}
  \label{fig:offset_show}
  \vspace{-10pt}
  \end{figure}

In the running scenario, the man's body sways up and down at each step while the man keeps moving forward. In this case, the motion is following a forward trend. In video prediction, we usually have to go through the whole frame sequence to get the motion trend. However, the future is unreachable. This dilemma is similar to reward prediction in reinforcement learning. Inspired by \textit{Temporal Difference} learning \cite{sutton1988learning}, we use an accumulating way to capture the pattern of motion variation. We use the previous motion filter $\mathcal{F}_{t-1}^l$ as the estimation of the current motion trend and get the momentum update function as follows:
\begin{equation}\label{equ:dynamic}
  \begin{split}
  & \mathcal{D}_{t}^l = \mathcal{D}_{t-1}^l + \alpha \left(\mathcal{F}_{t-1}^l - \mathcal{D}_{t-1}^l\right), \\
  \end{split}
  \vspace{-10pt}
\end{equation}
where $\alpha$ is the step size of momentum update and $\mathcal{D}_{t}^l$ is the learned trending momentum. We denote the above equation as $\mathcal{D}_{t}^l = \textrm{Trend}\left(\mathcal{F}_{t-1}^l, \mathcal{D}_{t-1}^l\right)$. With momentum update, $\mathcal{D}_{t}^l$ convergences to the weighted sum of motion filters $\mathcal{F}_{t}^l$, which can be viewed as the motion trend in the past period. In the running example (Figure~\ref{fig:offset_show}), $\mathcal{F}_{t-1}^l$ presents the motion of the last moment and $\mathcal{D}_{t}^l$ denotes the forward trend learned from the past. By momentum updating, this tendency estimation is of larger coefficient over time. Note that the trending momentum $\mathcal{D}_{t}^l$ is the momentum update of motion filter $\mathcal{F}_{t}^l$ and is also in the offset space, which presents the learned motion trend of pixels in a video. 

\subsubsection{Overall Procedure for MotionGRU} 

By implementing the key observation of motion decomposition, we design MotionGRU as the following procedure:
\begin{equation}
  \begin{split}
  \mathcal{F}_{t}^\prime & = \textrm{Transient}\left(\mathcal{F}_{t-1}^l, \textrm{Enc}(\mathcal{H}_{t}^l)\right)\\
  \mathcal{D}_{t}^l & = \textrm{Trend}\left(\mathcal{F}_{t-1}^l, \mathcal{D}_{t-1}^l\right)\\
  \mathcal{F}_{t}^l & = \mathcal{F}_{t}^\prime + \mathcal{D}_{t}^l \\
  m_{t}^{l} & = \textrm{Broadcast}\left(\sigma(W_\textrm{hm}\ast \textrm{Enc}(\mathcal{H}_{t}^l))\right) \\
  \mathcal{H}_t^\prime & = m_{t}^{l} \odot \textrm{Warp}\left(\textrm{Enc}(\mathcal{H}_{t}^l), \mathcal{F}_{t}^l\right)\\
  g_{t} & = \sigma \left(W_{1\times 1}\ast \textrm{Concat}([\textrm{Dec}(\mathcal{H}_t^\prime ), \mathcal{H}_{t}^{l}])\right) \\
  \mathcal{X}_{t}^l & = g_{t}\odot \mathcal{H}_{t-1}^l + (1-g_{t})\odot \textrm{Dec}(\mathcal{H}_t^\prime),\\
  \end{split}
  \label{equ:mornn_unit}
\end{equation}
where $t$ denotes the time step and $l\in\{1, \cdots, L\}$ denotes the current layer, \textrm{Transient}$(\cdot)$ and \textrm{Trend}$(\cdot)$ present the transient-variation learner and trending-momentum updater respectively. $\mathcal{F}_{t}^\prime$ and $\mathcal{D}_{t}^l$ denote the transient variation and trending momentum of the current frame. Based on the observation of motion decomposition, the motion filter $\mathcal{F}_{t}^l$ is the combination of transient variation and trending momentum. $m_{t}^{l}$ is the mask for motion filter and \textrm{Broadcast}$(\cdot)$ means the broadcast operation with kernel $W_\text{hm}$ to keep tensor dimension consistent to $\mathcal{H}_{t}^\prime$.

For the state transition, we use the warp operation \cite{black1996robust, brox2004high} to map the pixels from the previous state to the position in the next state, which is widely used in different fields of video analysis, such as video style transfer \cite{chen2017coherent} and video restoration \cite{wang2019edvr}. Here \textrm{Warp}$(\cdot)$ denotes the warp operation with bilinear interpolation. As shown in Figure \ref{fig:offset_show}, warping the previous state by the learned motion $\mathcal{F}_{t}^l$, we can explicitly incorporate motion variation into the transition of hidden states. More details about the warp operation in MotionGRU can be found in the Appendix \ref{appendix:implementation}. As shown in Figure \ref{fig:mornn_unit}, the final output $\mathcal{X}_{t}^l$ of MotionGRU is a gate $g_t$ controlled result from the input $\mathcal{H}_{t}^l$ and the decoder output, in which the decoder output has been explicitly transited by warp operation based on the motion filter $\mathcal{F}_{t}^l$.

Overall, by capturing the transient variation and motion trend separately and fusing them in a unified unit, MotionGRU can effectively model the spacetime-varying motions. With MotionGRU and Motion Highway, our MotionRNN framework can be applied to scenarios with ever-changing motions, which seamlessly compensates existing models.

\section{Experiments}

We extensively evaluate our proposed MotionRNN on the following three challenging benchmarks.

\vspace{-8pt}
\paragraph{Human motions. } 
This benchmark is built on the Human3.6M \cite{ionescu2013human3} dataset, which contains human actions from real world of $17$ different scenarios with $3.6$ million poses. We resize each RGB frame to the resolution of $128 \times 128$. Real-world human motion is much more complicated. For example, when a person is walking, different parts of the human body will have diverse transient variations, \textit{e.g.} the arms and legs are bending, the body is swaying. The complex motion variations will make the prediction of real human motion a really challenging task. 

\vspace{-8pt}
\paragraph{Precipitation nowcasting. }
Precipitation nowcasting is a vital application of video prediction. It is challenging to predict the accumulation, deformation, dissipation, or diffusion of radar echos reflecting severe weather.
This benchmark uses the Shanghai radar dataset, which contains evolving radar maps from Shanghai weather bureau. The Shanghai dataset has $40,000$ consecutive radar observations, collected every $12$ minutes, with $36,000$ sequences for training and $4,000$ for testing. Each frame is resized to the resolution of $64\times 64$. 

\vspace{-8pt}
\paragraph{Varied moving digits. }
We introduce the Varied Moving MNIST (V-MNIST) dataset consisting of sequences of frames with a resolution of $64\times 64$. Previous Moving MNIST \cite{srivastava2015training} or Moving MNIST++ \cite{shi2017deep} digits move with a lower velocity without digits variations. By contrast, our varied Moving MNIST forces all digits to move, rotate, and scale simultaneously. The V-MNIST are generated on the fly by sampling two different MNIST digits, with $100,000$ sequences for training and $10,000$ for testing. 

\vspace{-10pt}
\paragraph{Backbone models. }
To verify the universality of MotionRNN, we use the following predictive models as our backbone models including ConvLSTM \cite{shi2015convolutional}, PredRNN \cite{wang2017predrnn}, MIM \cite{WangZZLWY19} and E3D-LSTM \cite{wang2019eidetic}. On all benchmarks, our MotionRNN based on these models has four stacked blocks with 64-channel hidden states.
For E3D-LSTM, we replace the encoder and decoder inside the MotionGRU with 3D convolutions to downsample the 3D feature map to 2D and keep the other operations unchanged. 

\vspace{-10pt}
\paragraph{Implementation details. }
Our method is trained with the $L1+L2$ loss \cite{wang2019eidetic} to enhance the sharpness and smoothness of the generated frames simultaneously, using the ADAM \cite{Kingma2014Adam} optimizer with an initial learning rate of $3\times 10^{-4}$. The momentum factor $\alpha$ is set to 0.5. For memory efficiency, the learned filter size of MotionGRU is set to $3\times3$. The batch size is set to $8$, and the training process is stopped after $100,000$ iterations. All experiments are implemented in PyTorch \cite{PaszkeGMLBCKLGA19} and conducted on NVIDIA TITAN-V GPUs.

\subsection{Human Motion}

\paragraph{Setups.} 
We follow the experimental setting in MIM \cite{WangZZLWY19}, which uses the previous 4 frames to generate the future 4 frames. As for evaluation metrics, we use the frame-wise structural similarity index measure (SSIM), the mean square error (MSE), the mean absolute error (MAE) to evaluate our models. Besides these common metrics, we also use the Fr\' echet Video Distance (FVD) \cite{unterthiner2018towards}, which is a metric for qualitative human judgment of generated videos. The FVD could measure both the temporal coherence of the video content and the quality of each frame.

\begin{table}
\small
\caption{Quantitative results of Human3.6M upon different network backbones ConvLSTM \cite{shi2015convolutional}, MIM \cite{WangZZLWY19}, PredRNN \cite{wang2017predrnn} and E3D-LSTM \cite{wang2019eidetic}. A lower MSE, MAE or FVD, or a higher SSIM indicates a better prediction.}
\label{human_compare}
\setlength{\tabcolsep}{5.5pt}
\vspace{-8pt}
\begin{center}
\begin{tabular}{|l|cccc|}
\hline
Method & SSIM & MSE/10 & MAE/100 & FVD \\
\hline\hline
TrajGRU \cite{shi2017deep} & 0.801 & 42.2 & 18.6 & 26.9 \\
Conv-TT-LSTM \cite{su2020convolutional} & 0.791 & 47.4 & 18.9 & 26.2 \\
\hline
ConvLSTM \cite{shi2015convolutional} & 0.776 & 50.4 & 18.9 & 28.4 \\
\textbf{+ MotionRNN} & 0.800 & 44.3 & 18.6 & 26.9 \\
\hline
MIM \cite{WangZZLWY19}& 0.790 & 42.9 & 17.8 & 21.8 \\
\textbf{+ MotionRNN} & 0.841 & 35.1 & 14.9 & 18.3 \\
\hline
PredRNN \cite{wang2017predrnn} & 0.781 & 48.4 & 18.9 & 24.7 \\
\textbf{+ MotionRNN} & 0.846 & \textbf{34.2} & \textbf{14.8} & \textbf{17.6} \\
\hline
E3D-LSTM \cite{wang2019eidetic} & 0.869 & 49.4 & 16.6 & 23.7 \\
\textbf{+ MotionRNN} & \textbf{0.881} & 44.5 & 15.8 & 21.7 \\
\hline
\end{tabular}
\end{center}
\vspace{-10pt}
\end{table}

\begin{figure}[tbp]
\vspace{-5pt}
\centering
\includegraphics[width=\columnwidth]{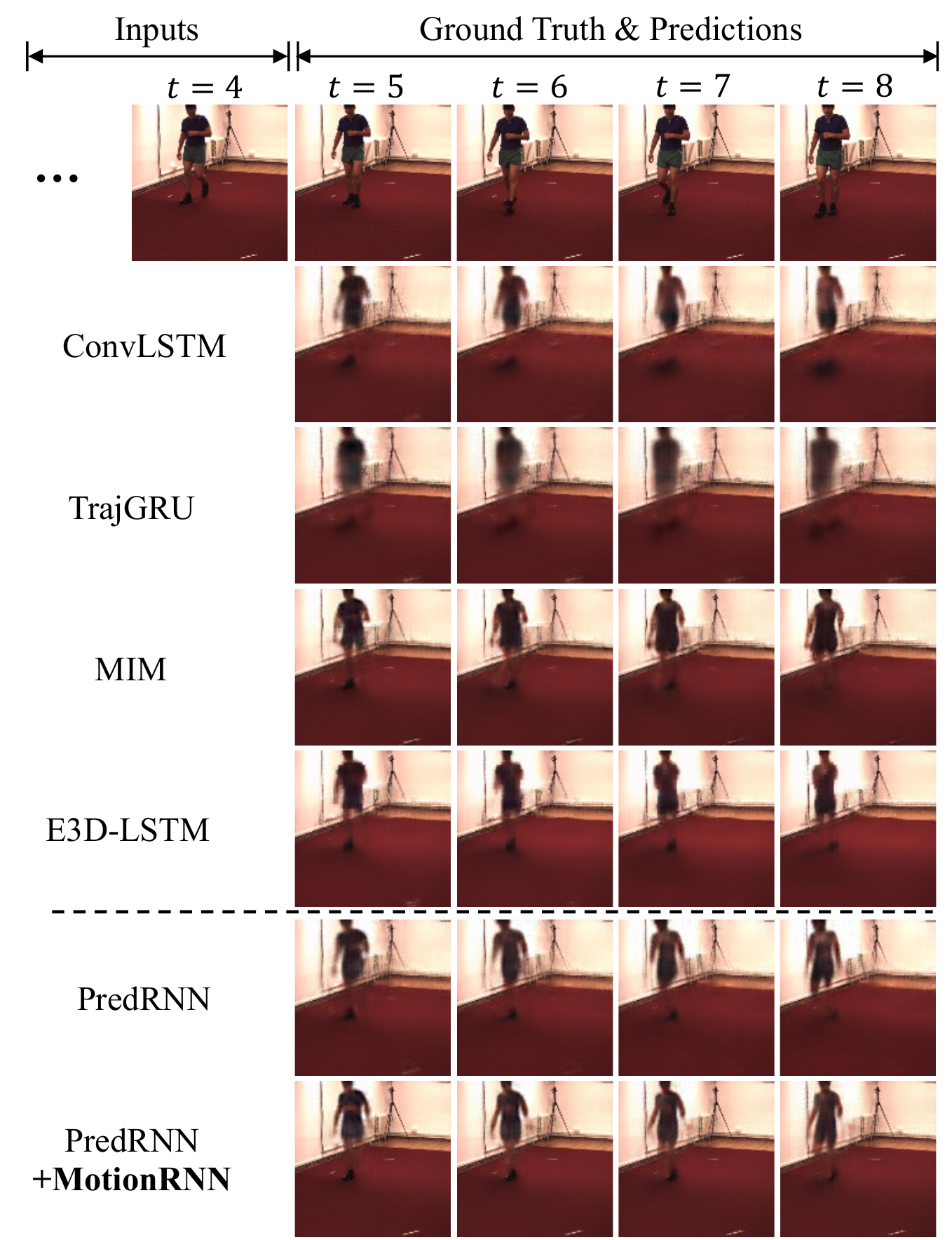}
\caption{Prediction frames on the human motion benchmark.}
\label{fig:human_exp}
\vspace{-15pt}
\end{figure}

\vspace{-10pt}
\paragraph{Results. }
As shown in Table \ref{human_compare}, our proposed MotionRNN promotes diverse backbone predictive models with consistent improvement in quantitative results. Significantly, with MotionRNN the performance improves \textbf{29\%} in MSE and \textbf{22\%} in MAE using the PredRNN as the backbone. Our approach also promotes the FVD, which means the prediction performs better in motion consistency and frame quality. To our best knowledge, MotionRNN based on PredRNN has achieved the \textbf{state-of-the-art} performance on Human3.6M. As for qualitative results, we show a case of walking in Figure \ref{fig:human_exp}. In this case, the human has a left movement tendency with transient variations across different body parts. The frames generated by MotionRNN are richer in detail and less blurry than those of other models, especially for the arms and legs. PredRNN and MIM may present the prediction in good sharpness but fail to bend the left elbow, and the predicted legs are also blurry. By contrast, MotionRNN could predict the sharpest sequence compared with previous methods and largely enrich the detail for each part of the body, especially for the arms and legs. What's more, the pose prediction for the arms and the legs is also predicted more precisely, which means our approach could not only maintain the details but perform well in motion capturing.

\begin{table}
\begin{center}
\caption{Parameters and computations comparison of MotionRNN using diverse backbone models. FLOPs denotes the number of multiplication operations for a human sequence prediction, which predicts the future 4 frames based on the previous 4 frames.}
\setlength{\tabcolsep}{5.4pt}
\label{human_memory_compare}
\begin{tabular}{|l|cc|c|}
\hline
Method & Params(MB) & FLOPs(G) & MSE$\Delta$\\
\hline\hline
ConvLSTM & 4.41 & 31.6 & - \\
\textbf{+ MotionRNN} & 5.21($\uparrow$ 18\%) & 36.6($\uparrow$ 16\%) & 12\% \\
\hline
PredRNN & 6.41 & 46.0 & - \\
\textbf{+ MotionRNN} & 7.01($\uparrow$ 9.3\%) & 49.5($\uparrow$ 7.6\%) & 29\% \\
\hline
MIM & 9.79 & 70.2 & - \\
\textbf{+ MotionRNN} & 10.4($\uparrow$ 6.2\%) & 73.7($\uparrow$ 5.0\%) & 18\% \\
\hline
E3D-LSTM & 20.4 & 292 & - \\
\textbf{+ MotionRNN} & 21.3($\uparrow$ 4.4\%) & 303($\uparrow$ 3.8\%) & 10\% \\
\hline
\end{tabular}
\end{center}
\vspace{-5pt}
\vspace{-5pt}
\end{table}

\vspace{-10pt}
\paragraph{Parameters and computations analysis.} 
We measure the complexity in terms of both model size and computations, as shown in Table \ref{human_memory_compare}. MotionRNN improves the performance of the PredRNN significantly (MSE: 48.4 $\rightarrow$ 34.2, SSIM: 0.781 $\rightarrow$ 0.846) with only 9.3\% additional parameters and 7.6\% increased computations. 
The increase of the model size is the same among different predictive frameworks because MotionRNN is only used as an external operator for hidden states across layers. The growth of the computations is also controllable. Based on these observations, we can see that our MotionRNN is a flexible model, which can improve the performance significantly on spatiotemporal variation modeling without significant sacrifice in model size or computation cost.

\begin{table}
\small
\begin{center}
\caption{The ablation of MotionRNN with respect to Motion Highway (\textbf{MH}), Transient Variation (\textbf{TV}) and Trending Momentum (\textbf{TM}) on the Human3.6M dataset. $\Delta$ denotes the MSE improvements over PredRNN.}\label{human_ablation_compare}
\setlength{\tabcolsep}{4.5pt}
\begin{tabular}{|l|ccc|c|c|}
\hline
Method & MH & TV & TM & $\frac{\textrm{MSE}}{10}$ & $\Delta$ \\
\hline\hline
PredRNN & & & & 48.4 & - \\
{+ Motion Highway} & $\surd$ & & & 42.5 & 12\% \\
{+ MotionGRU w/o Momentum} & & $\surd$ & & 41.5 & 14\% \\
{+ MotionGRU w/o Transient} & & &$\surd$ & 43.5 & 10\% \\
\hline
{+ MotionGRU} & & $\surd$ & $\surd$ & 40.3 & 17\% \\
{+ MotionRNN w/o Momentum} & $\surd$ & $\surd$ & & 38.9 & 20\% \\
{+ MotionRNN w/o Transient} & $\surd$ & & $\surd$ & 40.6 & 16\% \\
\hline
\textbf{+ MotionRNN} &$\surd$ & $\surd$ & $\surd$ & \textbf{34.2} & \textbf{29\%} \\
\hline
\end{tabular}
\vspace{-18pt}
\end{center}
\end{table}

\begin{figure}[htb]
\centering
\vspace{-5pt}
\includegraphics[width=0.85\columnwidth]{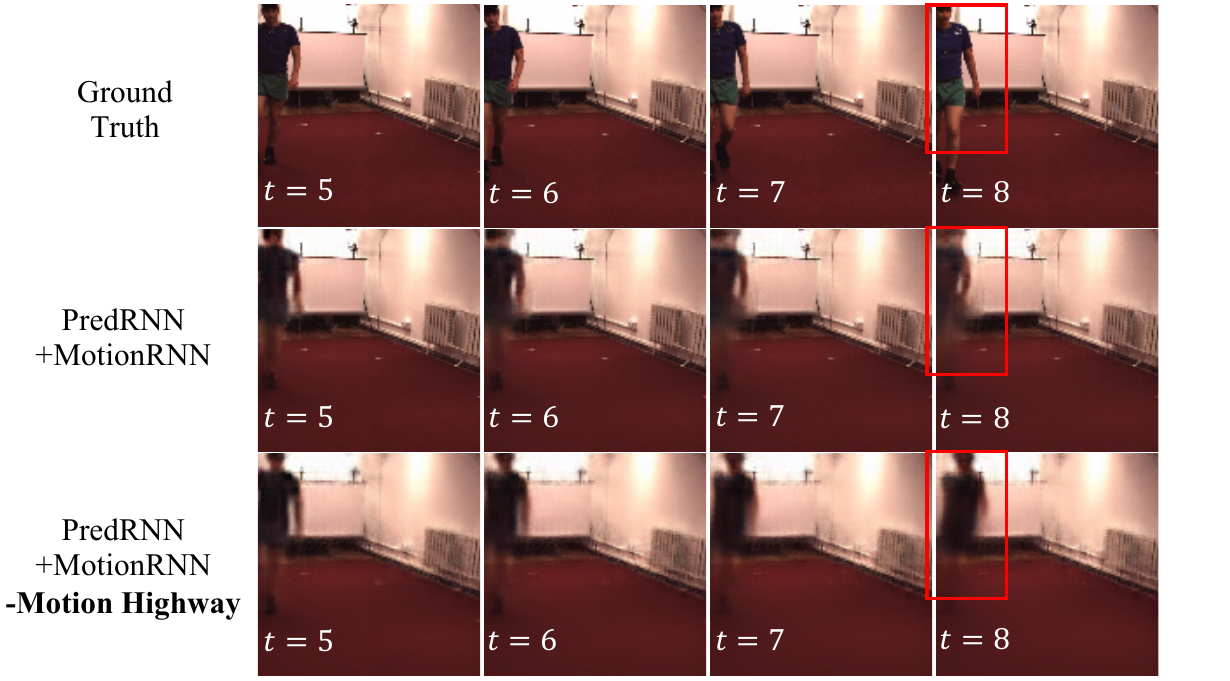}
\caption{The qualitative case for the ablation study of the Motion Highway, using the \textcolor{red}{red} box to box out of the body.}
\label{fig:human_mc}
\vspace{-5pt}
\end{figure}

\vspace{-10pt}
\paragraph{Ablation study.} 
As shown in Table \ref{human_ablation_compare}, we analyze the effectiveness of each part from our MotionRNN. Only by adopting the Motion Highway we could get a fairly good promotion (12\%$\uparrow$) indicating the Motion Highway can maintain the information of the motion context and compensate existing models for the additional useful information. 
Only adopting the MotionGRU without Motion Highway makes the MotionRNN achieve 17\% improvement. From the quantitative results described in Table \ref{human_ablation_compare}, we could easily find that the Motion Highway and MotionGRU can promote each other and achieve better improvement (\textbf{29\%$\uparrow$}). Furthermore, from the qualitative results shown in Figure \ref{fig:human_mc}, we can find without the Motion Highway, the predictions lose details of the arms and have the positional skewing. Thus we can verify the effect of our Motion Highway, which can compensate necessary content details to the MotionGRU and constrained the motion in the right area. More visualization can be found in 
Appendix \ref{appendix:Visualization_ablation}.
Besides, the learned trending momentum and transient variation give 9\% and 13\% extra promotions respectively. These results indicate that both parts of motion decomposition are effective for video prediction.

\begin{figure}[htb]
\vspace{-5pt}
\centering
\includegraphics[width=0.8\columnwidth]{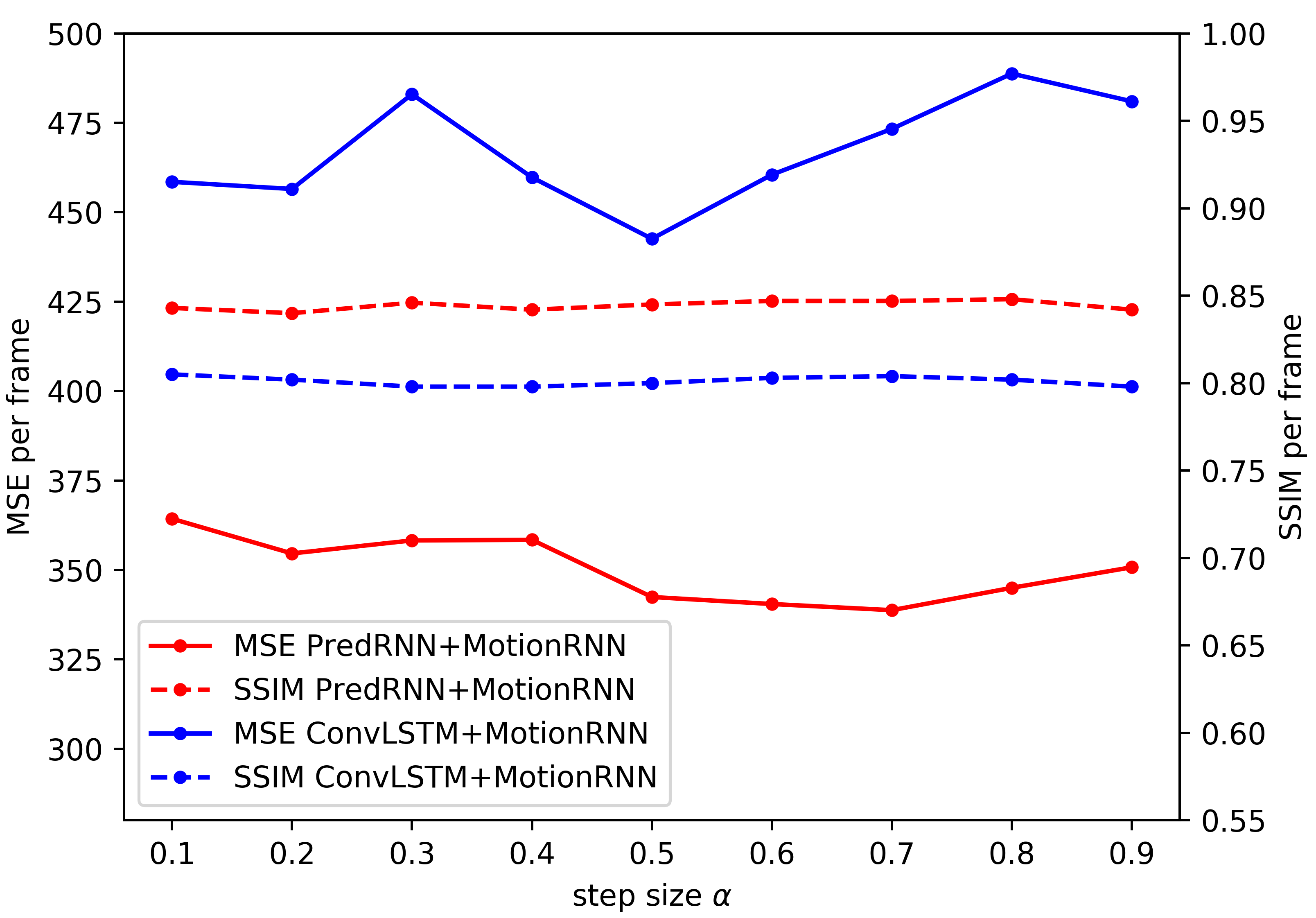}
\caption{The sensitivity analysis of hyper-parameter $\alpha$.}
\label{fig:human_alpha}
\vspace{-10pt}
\end{figure}

\vspace{-10pt}
\paragraph{Hyper-parameters.}
We show the sensitivity analysis of the training hyper-parameter $\alpha$ for trending momentum in Figure \ref{fig:human_alpha}. Our MotionRNN based on PredRNN and ConvLSTM achieves great performance when $\alpha=0.5$ and is robust and easy to tune in the range of $0.5$ to $0.7$. We have similar results on the other two benchmarks and thus set $\alpha$ to $0.5$ throughout the experiments.

\begin{table}
\small
\caption{Quantitative  results  of the Shanghai dataset upon different network backbone. A lower GDL or a higher CSI means a better prediction performance.}
\label{rader_compare}
\vspace{-10pt}
\begin{center}
\begin{tabular}{|l|ccccc|}
\hline
Method & SSIM & GDL & CSI30 & CSI40 & CSI50 \\
\hline\hline                                        TrajGRU & 0.815 & 13.9 & 0.576 & 0.545 & 0.484  \\
Conv-TT-LSTM & 0.820 & 13.6 & 0.571 & 0.530 & 0.469 \\
\hline
ConvLSTM & 0.837 & 12.3 & 0.624 & 0.605 & 0.560 \\
\textbf{+ MotionRNN} & 0.850 & 11.9 & 0.646 & 0.629 & 0.586 \\
\hline
MIM & 0.849 & 11.3  & 0.654 & 0.646 & 0.609 \\
\textbf{+ MotionRNN} & 0.863 & 11.1 & 0.668 & 0.654 & 0.614 \\
\hline
PredRNN & 0.841 & 11.9 & 0.633 & 0.622 & 0.581 \\
\textbf{+ MotionRNN} & 0.865 & 10.9 & \textbf{0.678} & \textbf{0.664} & \textbf{0.623} \\
\hline
E3D-LSTM & 0.842 & 12.7 & 0.615 & 0.615 & 0.590\\
\textbf{+ MotionRNN} & \textbf{0.880} & \textbf{9.67} & 0.671 & 0.659 & 0.621 \\
\hline
\end{tabular}
\end{center}
\vspace{-10pt}
\end{table}

\begin{figure}[tbp]
\vspace{-5pt}
\centering
\includegraphics[width=\columnwidth]{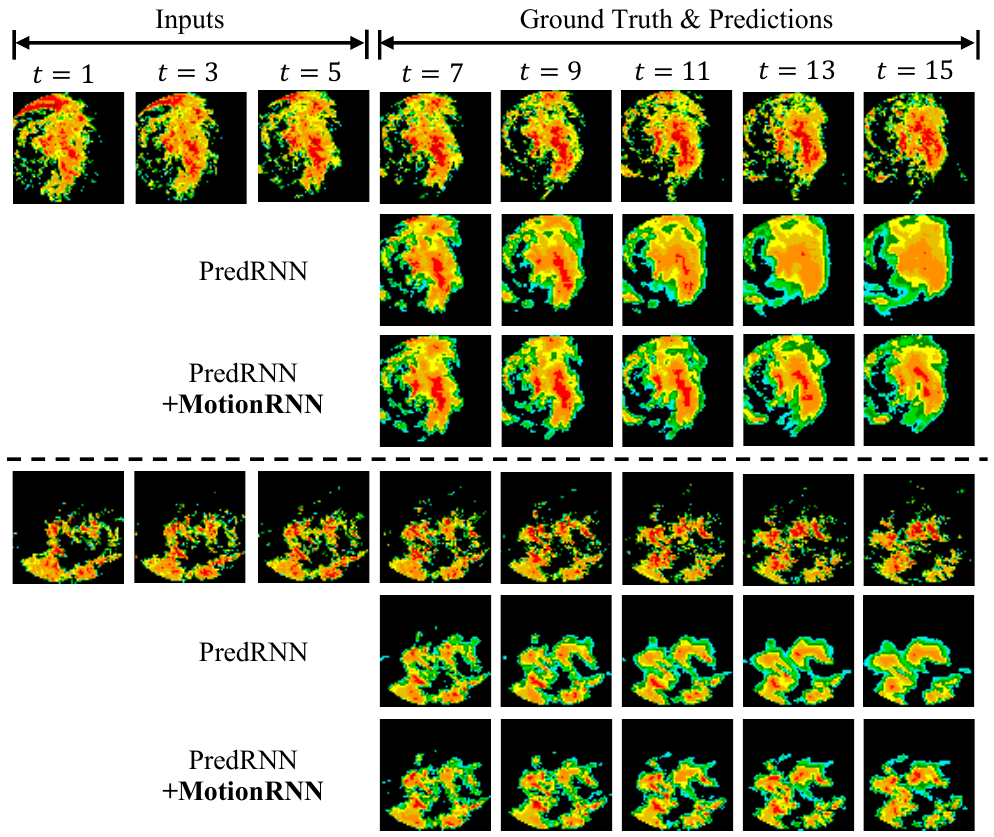}
\caption{Prediction examples on the Shanghai radar echo dataset. 
}
\label{fig:radar_exp}
\vspace{-10pt}
\end{figure}

\subsection{Precipitation Nowcasting}

\paragraph{Setups. } 
We forecast the next 10 radar echo frames from the previous 5 observations, covering weather conditions in the next two hours. We use the gradient difference loss (GDL) \cite{Mathieu2015Deep} to measure the sharpness of the prediction frames. A lower GDL indicates a higher sharpness similarity of ground truth. Further, for radar echo intensities, we convert the pixel values in dBZ and compare the Critical Success Index (CSI) with 30 dBZ, 40 dBZ, 50 dBZ as thresholds, respectively. CSI is defined as $\textrm{CSI}=\frac{\textrm{Hits}}{\textrm{Hits}+\textrm{Misses}+\textrm{FalseAlarms}}$, where hits correspond to the true positive, misses correspond to the false positive, and false alarms correspond to the false negative. A higher CSI indicates better forecasting performance. Compared with MSE, the CSI metric is particularly sensitive to the high-intensity echoes, always with high changeable motions.

\vspace{-10pt}
\paragraph{Results.} 
We provide quantitative results in Table \ref{rader_compare}, our MotionRNN using the state-of-the-art model E3D-LSTM achieves \textbf{24}\% improvement on the GDL metric, indicating our MotionRNN could produce the most sharpness predicted sequence.  With our MotionRNN, the predictive frameworks could significantly improve various CSI metrics with different thresholds, which demonstrates that our approach can make predictions well on the changeable radar echos. As shown in Figure \ref{fig:radar_exp}, MotionRNN predicts the motion more precisely in qualitative results. In the top case, there is a cyclone in which the motion contains moving up and anticlockwise rotation. PredRNN could roughly predict the cyclone positions but suffers from blurring. Focusing on the center part of prediction at $t=15$, MotionRNN forecasts the exact rotation trend, but the echoes predicted from PredRNN is just a block without cyclone rotation details. In the bottom case, the echoes have an upward-diffusion movement and a slight anticlockwise rotation simultaneously. Our approach provides more details for the diffusion than PredRNN. We find many small cloud clusters in the right-bottom area and a more subtle outline of the center part. We believe such an accurate, fine-grained prediction will be very valuable for severe weather forecasting.

\begin{figure}[htb]
\vspace{-5pt}
\centering
\includegraphics[width=\columnwidth]{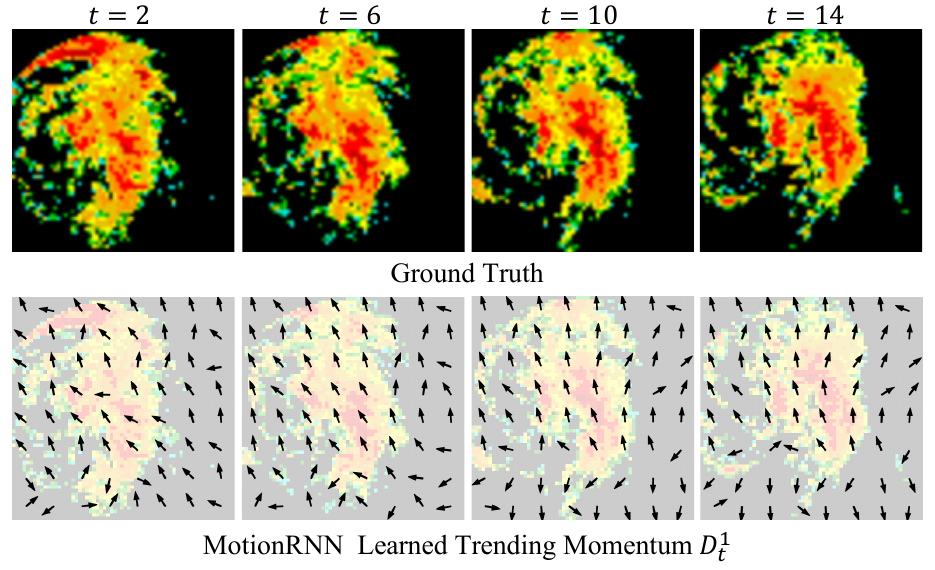}
\caption{The visualization of learned motion tendency, which shows the motion direction. The arrows are calculated from $\mathcal{D}_t^1$.
}
\label{fig:radar_show}
\vspace{-5pt}
\end{figure}

\vspace{-10pt}
\paragraph{Motion trend visualization.} 
To have a better view of the learned motion trend, we visualized the $\mathcal{D}_{t}^1$, the trending momentum of the first layer MotionGRU. 
We use the arrows to show the direction of the offsets, which represent the motion trend. 
The details about visualization operations are shown in the Appendix \ref{appendix:Visualization_detail}. In Figure \ref{fig:radar_show}, the center arrows show a moving up and anticlockwise rotation. The bottom arrows indicate the downward-motion of a cyclone's small tile. This visualization exactly shows that MotionRNN could capture the motion trend and have the ability to model the spacetime-varying motion in radar.

\subsection{Varied Moving Digits}

\paragraph{Setups. } 
We predict the future 10 frames based on the previous 10 frames. We use MSE, SSIM, GDL, and Peak Signal to Noise Ratio (PSNR) as evaluation metrics. Compared with the original Moving MNIST++ dataset, the sequences in our proposed varied Moving MNIST (V-MNIST) dataset are in lots of variations, such as faster moving speed, higher speed rotation, and scaling.

\begin{table}
\caption{Quantitative results of V-MNIST. Higher PSNR means better prediction. }
\vspace{-5pt}
\label{mnist_compare_table}
\begin{center}
\setlength{\tabcolsep}{8pt}
\begin{tabular}{|l|cccc|}
\hline
Method & MSE & SSIM & PSNR & GDL \\
\hline\hline
TrajGRU & 109 & 0.515 & 15.9 & 69.3 \\
Conv-TT-LSTM & 71.1 & 0.744 & 18.4 & 53.6  \\
\hline
E3D-LSTM & 57.6 & 0.852 & 19.7 & 44.6 \\
\textbf{+ MotionRNN} & 52.8 & 0.867 & 20.3 & 42.4 \\

\hline
ConvLSTM & 47.0 & 0.845 & 20.6 & 41.8 \\
\textbf{+ MotionRNN} & 44.4 & 0.861 & 20.9 & 40.3 \\

\hline
MIM & 34.6 & 0.888 & 22.3 & 34.6 \\
\textbf{+ MotionRNN} & 28.9 & 0.906 & 23.1 & 30.9  \\

\hline
PredRNN & 35.6 & 0.891 & 22.1 & 34.7  \\
\textbf{+ MotionRNN} & \textbf{25.1} & \textbf{0.920} & \textbf{24.0} & \textbf{27.7}  \\

\hline
\end{tabular}
\end{center}
\end{table}

\begin{figure}[tbp]
\vspace{-10pt}
\centering
\includegraphics[width=\columnwidth]{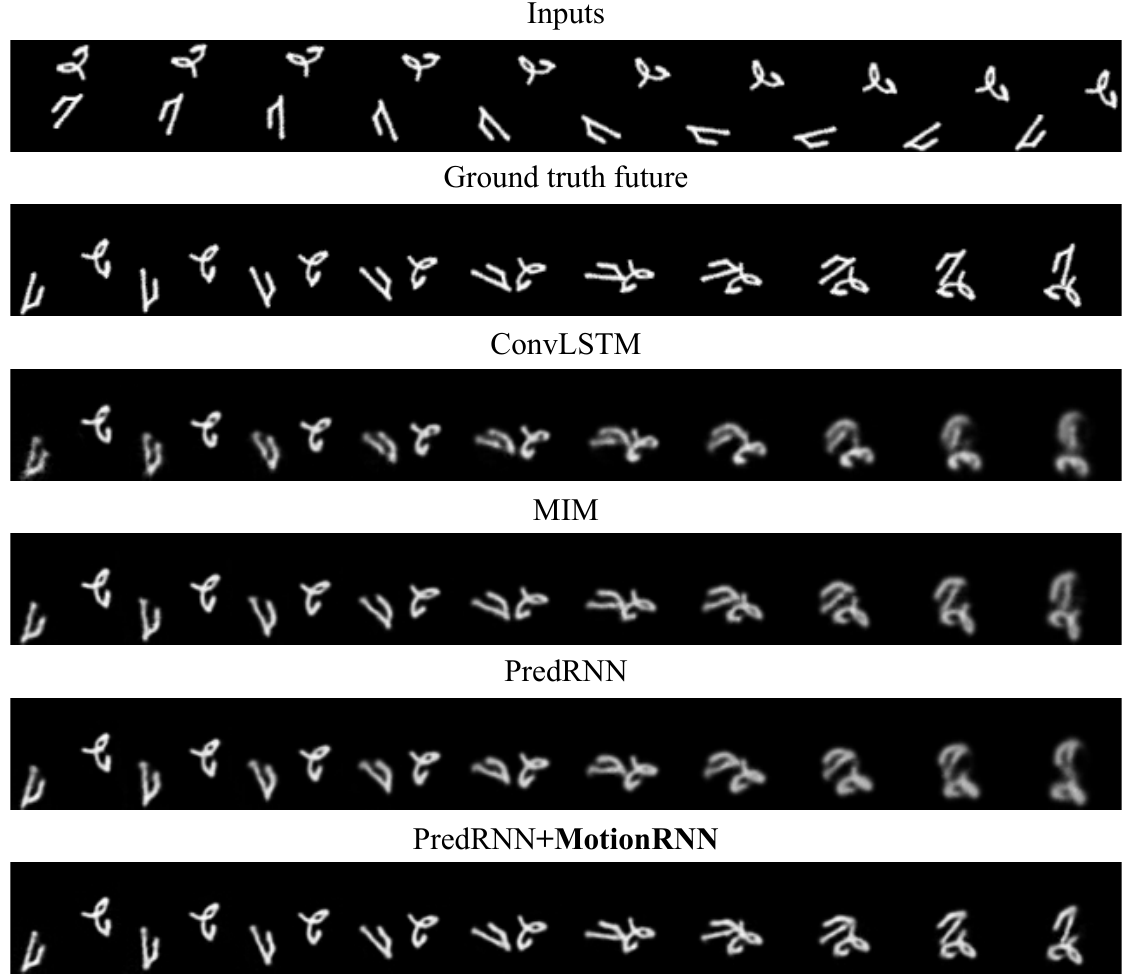}
\caption{Prediction examples of V-MNIST.}
\label{fig:mnist_exp}
\vspace{-10pt}
\end{figure}

\vspace{-10pt}
\paragraph{Results. }
With MotionRNN, the backbone models could get a consistent improvement in all metrics, as presented in Table \ref{mnist_compare_table}. Especially in PredRNN, the predictions gain an excellent promotion in MSE and GDL. In the case of Figure \ref{fig:mnist_exp}, the digits move with rotation, which makes the prediction task harder. Previous models fail in giving the prediction with enough sharpness and clear strokes. As shown in the bottom line, with MotionRNN, PredRNN presents more satisfactory and sharper results.

\section{Conclusion}

In this paper, we have presented a flexible MotionRNN framework to predict spacetime-varying motions. Based on the observation that the motion can be decomposed to transient variation and the motion trend, we design the MotionGRU to capture the transient variation of the motion and the motion tendency respectively. By incorporating the MotionGRU to RNN-based predictive frameworks with the motion highway, MotionRNN can model the motion explicitly in state transitions and avoid the motion vanishing.
With high flexibility, we apply MotionRNN with a series of predictive models to achieve significant promotions and state-of-the-art performance on three challenging prediction tasks.

\section*{Acknowledgments}

This work was supported by National Key R\&D Program of China (2017YFC1502003), NSFC grants (62022050, 62021002, 61772299, 71690231), Beijing Nova Program (Z201100006820041), and MOE Innovation Plan of China.

{
\bibliographystyle{ieee_fullname}
\bibliography{egbib}
}

\clearpage

\appendix

\section{MotionGRU: Implementation Details}\label{appendix:implementation}
In this section, we will give a detailed description of the implementation details for MotionGRU from the \textbf{tensor view}. We first recall the equations of MotionGRU:
\begin{equation}
  \begin{split}
  \mathcal{F}_{t}^\prime & = \textrm{Transient}\Big(\mathcal{F}_{t-1}^l, \textrm{Enc}(\mathcal{H}_{t}^l)\Big)\\
  \mathcal{D}_{t}^l & = \textrm{Trend}\Big(\mathcal{F}_{t-1}^l, \mathcal{D}_{t-1}^l\Big)\\
  \mathcal{F}_{t}^l & = \mathcal{F}_{t}^\prime + \mathcal{D}_{t}^l \\
  m_{t}^{l} & = \textrm{Broadcast}\Big(\sigma(W_\textrm{hm}\ast \textrm{Enc}(\mathcal{H}_{t}^l))\Big) \\
  \mathcal{H}_t^\prime & = m_{t}^{l} \odot \textrm{Warp}\Big(\textrm{Enc}(\mathcal{H}_{t}^l), \mathcal{F}_{t}^l\Big)\\
  g_{t} & = \sigma \Big(W_{1\times 1}\ast \textrm{Concat}([\textrm{Dec}(\mathcal{H}_t^\prime ), \mathcal{H}_{t}^{l}])\Big) \\
  \mathcal{X}_{t}^l & = g_{t}\odot \mathcal{H}_{t-1}^l + (1-g_{t})\odot \textrm{Dec}(\mathcal{H}_t^\prime),\\
  \end{split}
  \label{equ:mornn_unit_appendix}
\end{equation}
where subscript $t$ denotes the time step, the superscript $l\in\{1, \cdots, L\}$ denotes the current layer, $\sigma$ donates the sigmoid function, $\ast$ and $\odot$ denote the convolution and the Hadamard product respectively.
\vspace{-5pt}
\paragraph{Learned Motion. } 
$\mathcal{H}_{t}^{l}\in \mathbb{R}^{C\times H\times W}$ donates the hidden state of original predictive models. For memory efficiency, we employ the Encoder-Decoder structure, the encoder use the stride 2 convolution and $\mathrm{Enc}(\mathcal{H}_{t}^{l})\in \mathbb{R}^{\frac{C}{4}\times\frac{H}{2}\frac{W}{2}}$. $\mathcal{F}_{t}^\prime$ and $\mathcal{D}_{t}^{l}$ present the learn transient variation and trending momentum respectively, which present the pixel-wise offsets of the state. Here, $\mathcal{F}_{t}^\prime,\mathcal{D}_{t}^{l}\in\mathbb{R}^{2k^2\times\frac{H}{2}\times\frac{W}{2}}$, where $k$ is the filter size. $(\mathcal{F}_{t}^\prime)_{:,m,n}$ is a 1-dim tensor with size $2k^2$ and represent the vertical and horizontal offsets of the pixel at $(m,n)$ position and the adjacent area with size $k^2$. Thus, the motion filter $\mathcal{F}_{t}^l\in\mathbb{R}^{2k^2\times\frac{H}{2}\times\frac{W}{2}}$ donates the learned motion-based state transition. With kernel $W_{hm}$, the channel number of the tensor is change to $k^2$. The \textrm{Broadcast} means the operation to broadcast and transpose the $k^2\times \frac{H}{2}\times \frac{W}{2}$ tensor to $\frac{C}{4}\times \frac{H}{2}\times \frac{W}{2}\times k^2$. $m_{t}^{l}\in \mathbb{R}^{\frac{C}{4}\times \frac{H}{2}\times \frac{W}{2}\times k^2}$ is the mask for the motion of the $k \times k$ filter area. 
\vspace{-5pt}
\paragraph{Warp. } 
The \textrm{Warp} donates the warp operation with bilinear interpolation in a unit square. It can select the pixels in $\text{Enc}(\mathcal{H}_{t}^l)$, which point out to the $k\times k$ filter area by offset $\mathcal{F}_{t}^l$. It can be formulated as follows:
\begin{equation}\label{equ:mornn_unit_warp}
  \begin{split}
  & \text{Warp}(\mathcal{H}, \mathcal{F}) = \Big(\mathcal{H}_{c,m,n,i}^\prime\Big)_{\frac{C}{4}\times \frac{H}{2}\times \frac{W}{2}\times k^2}  \\  
  & = \Big(\mathcal{H}_{c,(m+p_{ix}-\mathcal{F}_{i,m,n}),(n+p_{iy}-\mathcal{F}_{i+k^2,m,n})}\Big)_{\frac{C}{4}\times \frac{H}{2}\times \frac{W}{2}\times k^2}, \\ 
  \end{split}
\end{equation}
where $i\in\{1,2,\cdots,k^2\}$, $p_{ix}=[\frac{i}{k}]-[\frac{k}{2}]$, $p_{iy}=(i\ \text{mod}\ k)-[\frac{k}{2}]$. 
The value of select pixel is calculated by the bilinear interpolation in the unit square as follows:
\begin{equation}
\begin{split}
& \mathcal{H}_{c,m^\prime,n^\prime} \\
&=
\begin{bmatrix}
1-m^{\prime\prime} & m^{\prime\prime}
\end{bmatrix}
\begin{bmatrix}
\mathcal{H}_{c,\lfloor m^\prime \rfloor,\lfloor n^\prime \rfloor} & \mathcal{H}_{c,\lfloor m^\prime \rfloor,\lceil n^\prime \rceil} \\
\mathcal{H}_{c,\lceil m^\prime \rceil,\lfloor n^\prime \rfloor} & \mathcal{H}_{c,\lceil m^\prime \rceil,\lceil n^\prime \rceil} \\
\end{bmatrix}
\begin{bmatrix}
1-n^{\prime\prime} \\
n^{\prime\prime}
\end{bmatrix}
\end{split}
\end{equation}
where $\lfloor\cdot\rfloor$ and $\lceil\cdot\rceil$ donate the floor function and ceiling function respectively, $m^{\prime\prime}=m^\prime-\lfloor m^\prime\rfloor$, $n^{\prime\prime}=n^\prime-\lfloor n^\prime\rfloor$.
\vspace{-5pt}
\paragraph{Gated Output. } 
$\mathcal{H}_{t}^\prime\in \mathbb{R}^{\frac{C}{4}\times \frac{H}{2}\times \frac{W}{2}\times k^2}$ donates the wrapped feature map. The decoder squeezes $\mathcal{H}_{t}^\prime$ to $\mathbb{R}^{\frac{C}{4}\times \frac{H}{2}\times \frac{W}{2}}$ by $1\times 1$ convolution and deconvolution to $\mathbb{R}^{C\times H\times W}$. The output gate $g_{t}\in \mathbb{R}^{C\times H\times W}$ is calculated from the concatenation of the input $\mathcal{H}_{t}^l$ and the decoded feature $\mathrm{Dec}(\mathcal{H}_{t}^\prime)$ by $1\times1$ convolution $W_{1\times1}$. The output $\mathcal{X}_{t}^{l}\in\mathbb{R}^{C\times H\times W}$ has the same shape as $\mathcal{H}_{t}^{l}$ and presents the motion-based transited state.

\section{Visualization of Ablation Study} \label{appendix:Visualization_ablation}
\vspace{-5pt}
In addition to the intutive showcase in Figure \ref{fig:human_mc}, we will provide more visualization for the ablation study to make each part's effect of MotionRNN more comprehensible.

Figure \ref{fig:visual_ablation_human} visualizes more ablation cases.
Without transient variation, the predictions lose pivotal details of hand and leg movement. Without trending momentum, the model fails to predict the human's back position correctly.

\begin{figure}[hbp]
\centering
\vspace{-5pt}
\animategraphics[width=\columnwidth]{4}{pic/human_gif_new_new-}{1}{10}
\caption{Visualization of ablation study. \textbf{Click to see the video}. Adobe reader is required to view the animation.}
\label{fig:visual_ablation_human}
\vspace{-15pt}
\end{figure}

\section{The Detail of Motion Trend Visualization}\label{appendix:Visualization_detail}
\vspace{-5pt}
In this section, we will detail the visualization of learned trending momentum $\mathcal{D}_{t}^{l}$ corresponding to Figure \ref{fig:radar_show}, which indicates the learned motion trend.

Take the Radar Shanghai dataset as an example. The input frame is with the resolution of $64\times 64\times 1$. Hidden states have 64 channels. After the patch operation with 4 patch number and the encoder with stride 2 inside the MotionGRU, the hidden state is converted to $\mathrm{Enc}(H)_{t}^l\in\mathbb{R}^{16\times 8\times 8}$. $\mathcal{D}_{t}^{l}\in\mathbb{R}^{18\times 8\times 8}$ can be 
split into two tensors. Each of them is in $\mathbb{R}^{9\times 8\times 8}$ and presents the learned offsets of $3\times 3$ filter area in vertical and horizontal respectively. The learned offsets just present the motion. For visualization, we first calculate the mean along the channel dimension and get the x-offset and y-offset of every pixel of the $\mathrm{Enc}(\mathcal{H}_{t}^l)$. After the length regularization, we use the arrows to show the direction of the offsets, which represents the motion tendency.

\begin{figure}[hbp]
\centering
\vspace{-5pt}
\includegraphics[width=\columnwidth]{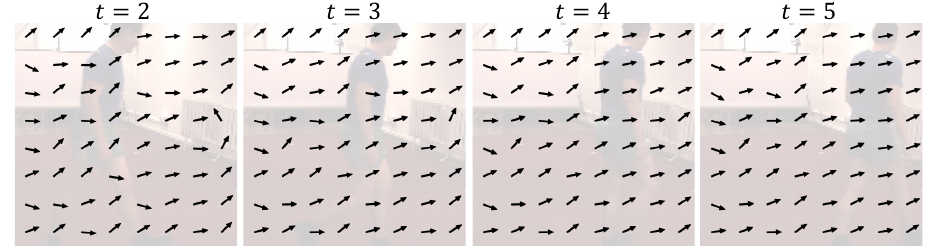}
\caption{Visualization of learned motion trend in human case.}
\label{fig:visual_trend_human}
\vspace{-15pt}
\end{figure}

Following the above method, we can also visualize the learned motion trend of the human case in Figure \ref{fig:human_show}. As shown in Figure \ref{fig:visual_trend_human}, the learned motion trend indicates a left-to-right tendency, which is just the global movement.

\section{Guangzhou Benchmark}
\vspace{-5pt}
Besides the radar data from Shanghai, we also explore another challenging radar dataset \cite{WangZZLWY19}, which its radar echos are collected every 6 minutes, from May 1st, 2014 to June 30th, 2014 in Guangzhou. The Guangzhou dataset is full of rain, and it can be used as an example of weather forecasting in rainy areas.
\vspace{-10pt}
\paragraph{Setups.} 
We also follow the experimental setting MIM \cite{WangZZLWY19}, which has achieved the \textbf{state-of-the-art} performance in the Guangzhou dataset. We use the previous 10 frames to generate the future 10 frames. As for evaluation metrics, we use the CSI with 30 dBZ, 40 dBZ, 50 dBZ as thresholds.

\begin{table}
\small
\caption{Quantitative  results  of the Radar Echo dataset upon different network backbones. A  higher CSI means a better performance.}
\vspace{-15pt}
\label{rader_echo_compare}
\setlength{\tabcolsep}{14pt}
\begin{center}
\begin{tabular}{|l|ccc|}
\hline
Method & CSI30 & CSI40 & CSI50 \\
\hline\hline
TrajGRU\cite{shi2017deep}  & 0.251 & 0.201 & 0.150  \\
\hline
E3D-LSTM\cite{wang2019eidetic}  & 0.350 & 0.309 & 0.254 \\
\textbf{+ MotionRNN}  & 0.379  & 0.366 & 0.333 \\
\hline
ConvLSTM\cite{shi2015convolutional}   & 0.385 & 0.360 & 0.315 \\
\textbf{+ MotionRNN} & 0.411  & 0.387 & \textbf{0.350} \\
\hline
PredRNN\cite{wang2017predrnn}   & 0.401 & 0.378 & 0.306 \\
\textbf{+ MotionRNN}  & 0.424 & 0.391 & 0.345 \\
\hline
MIM\cite{WangZZLWY19}  & 0.429 & 0.399 & 0.317 \\
\textbf{+ MotionRNN}   & \textbf{0.431} & \textbf{0.403} & 0.343 \\
\hline
\end{tabular}
\vspace{-25pt}
\end{center}
\end{table}

\begin{figure}[hbp]
\centering
\vspace{-5pt}
\includegraphics[width=\columnwidth]{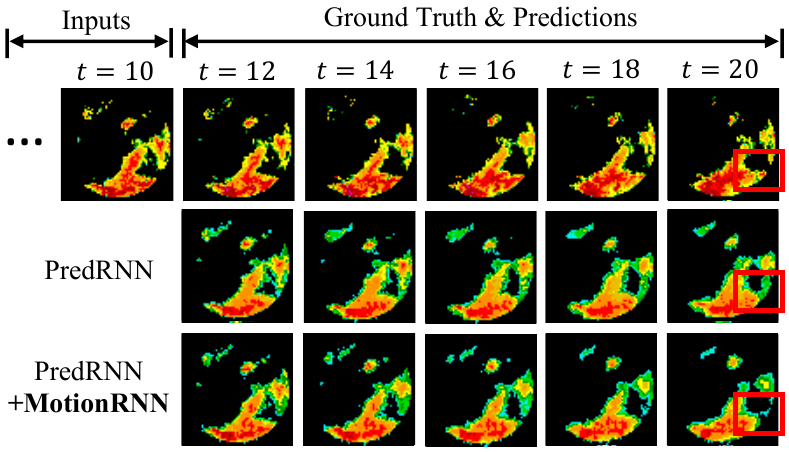}
\caption{Prediction examples on the Radar Echo. }
\label{fig:echo_exp_1}
\vspace{-15pt}
\end{figure}

\vspace{-10pt}
\paragraph{Results.} 
As shown in Table \ref{rader_echo_compare}, the predictive models can get a consistent improvement in CSI with different thresholds, even in the \textbf{state-of-the-art} method MIM. Especially, MotionRNN can significantly promote CSI50. It means we can enhance the forecasts of severe weather through MotionRNN. Furthermore, as shown in Figure \ref{fig:echo_exp_1}, the prediction results are more precise, indicating that the MotionRNN are better in detail prediction and more accurate on thick clouds prediction.

\section{More Qualitative Results}
\vspace{-5pt}
We will show more qualitative cases in this section. Areas to focus on are highlighted in \textcolor{red}{red}. 

\begin{figure}[hbp]
\centering
\vspace{-5pt}
\includegraphics[width=0.98\columnwidth]{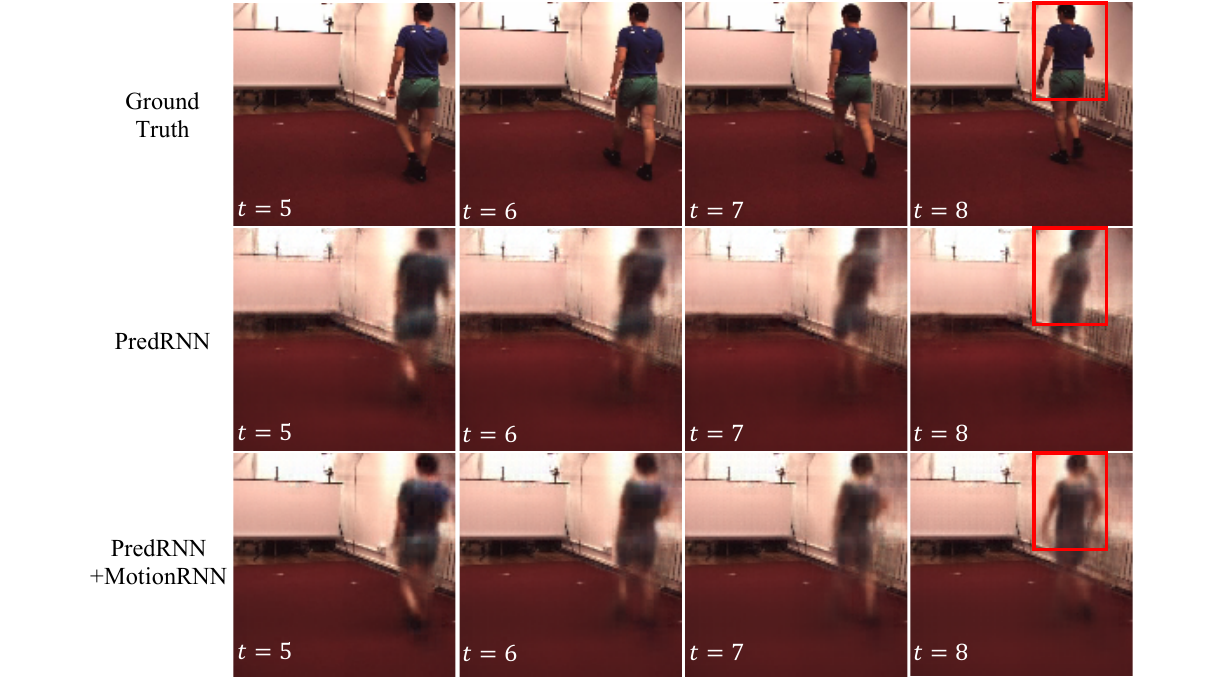}
\caption{Prediction examples on the Human3.6M. 
}
\label{fig:human_exp_1}
\vspace{-10pt}
\end{figure}

\begin{figure}[hbp]
\centering
\vspace{-10pt}
\includegraphics[width=0.92\columnwidth]{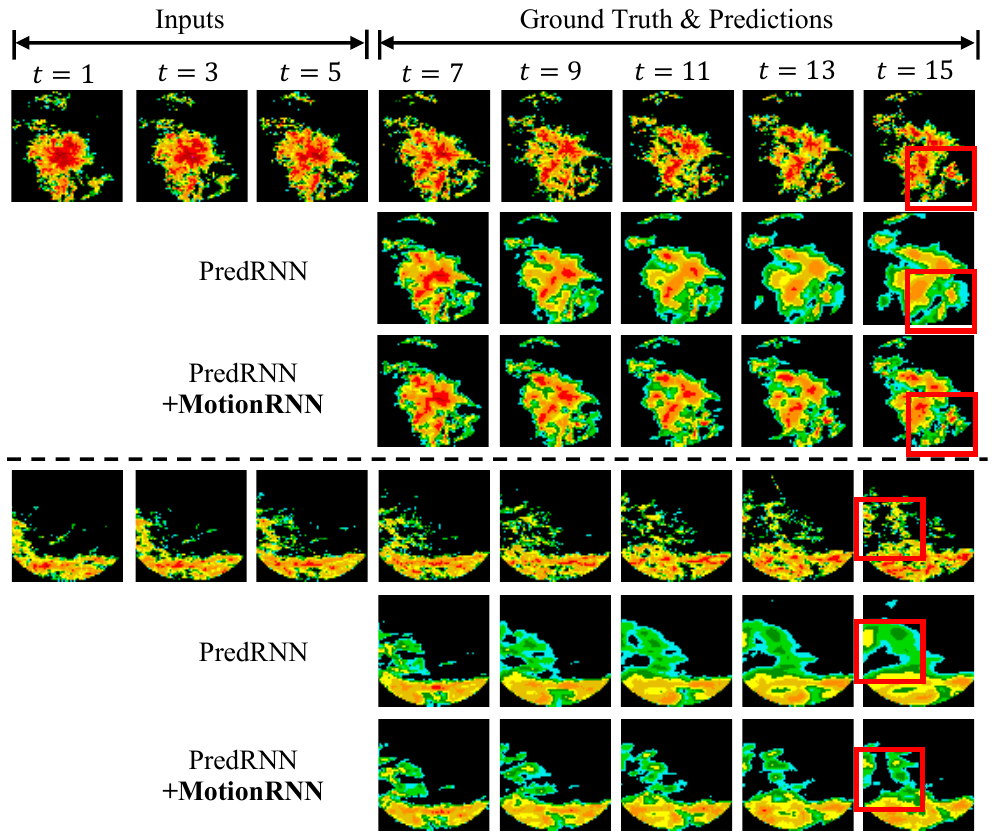}
\caption{Prediction examples on the Radar Shanghai. 
}
\label{fig:radar_external}
\end{figure}

\begin{figure}[hbp]
\centering
\vspace{-18pt}
\includegraphics[width=0.92\columnwidth]{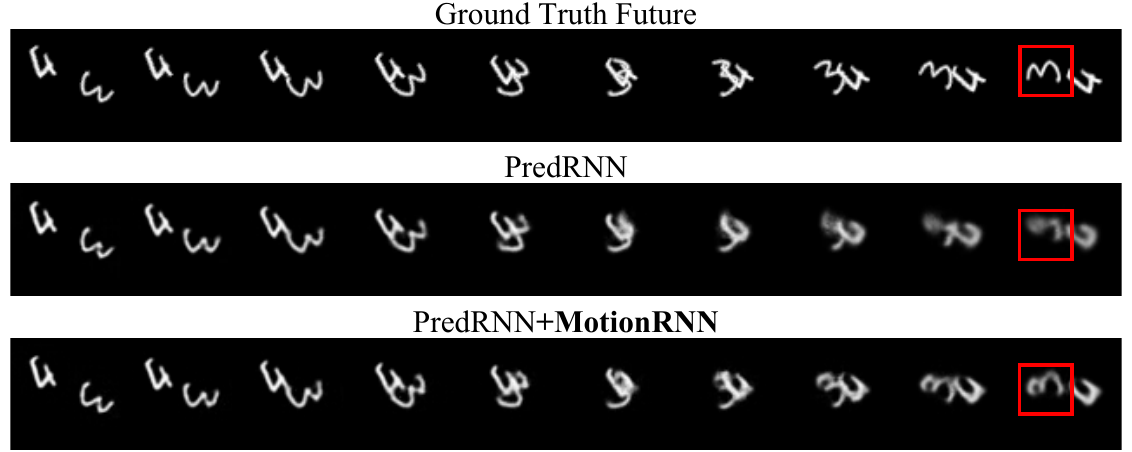}
\caption{Prediction examples on the V-MNIST. 
}
\label{fig:mnist_external}
\end{figure}

In human motion prediction (Figure \ref{fig:human_exp_1}), MotionRNN presents the details of human arms, while the predicted arms by PredRNN are vanished. In Figure \ref{fig:radar_external}, with MotionRNN, predictive models can generate the results with better sharpness. In synthetic V-MNSIT dataset (Figure \ref{fig:mnist_external}), our approach can generate eidetic results of number ``3". These qualitative results show that MotionRNN can significantly improve the accuracy and sharpness of the prediction.

\end{document}